\documentclass[journal]{IEEEtran}

\usepackage{cite}

\usepackage[pdftex]{graphicx}
\graphicspath{{./}}
\DeclareGraphicsExtensions{.jpg}
\usepackage{pdfpages}

\usepackage{amsmath}
\usepackage{amsfonts}
\usepackage{algorithmic}
\usepackage{array}
\usepackage{url}
\usepackage{animate}

\definecolor{citecolor}{RGB}{119,185,0} 
\usepackage[pagebackref=false,breaklinks=true,letterpaper=true,colorlinks,citecolor=citecolor,bookmarks=false]{hyperref}
\usepackage{soul}
\usepackage[utf8]{inputenc}
\usepackage{multirow}
\usepackage{comment}

\usepackage{pifont}
\def\ie{\emph{i.e.}} 
\def\etal{\emph{et~al. }} 

\newcommand{\zznote}[1]{\textcolor{red}{[ZZ: #1]}}

\newlength\savewidth\newcommand\shline{\noalign{\global\savewidth\arrayrulewidth
  \global\arrayrulewidth 1pt}\hline\noalign{\global\arrayrulewidth\savewidth}}
  
\begin{document}
\title{ Jointly Harnessing Prior Structures and Temporal Consistency for Sign Language Video Generation 
}
\author{Yucheng Suo, Zhedong Zheng, Xiaohan Wang, Bang Zhang and Yi Yang,~\IEEEmembership{Senior~Member,~IEEE} \thanks{
Yucheng Suo, Xiaohan Wang and Yi Yang are with the College of Computer Science and Technology, Zhejiang
University, China 310027. E-mail: suoych@gmail.com, xiaohan.wang@zju.edu.cn, yangyics@zju.edu.cn.}
\thanks{Zhedong Zheng is with Sea-NExT Joint Lab, the School of Computing, National University of Singapore, Singapore 118404. E-mail: zdzheng@nus.edu.sg.} 
\thanks{Bang Zhang is with DAMO Academy, Alibaba Group, China 311121. E-mail: zhangbang.zb@@alibaba-inc.com.} 
}

\markboth{Journal of \LaTeX\ Class Files,~Vol.~14, No.~8, August~2015}%
{Shell \MakeLowercase{\textit{et al.}}: Bare Demo of IEEEtran.cls for IEEE Journals}

\maketitle

\begin{abstract}
Sign language is the window for people differently-abled to express their feelings as well as emotions. However, it remains challenging for people to learn sign language in a short time. To address this real-world challenge, in this work, we study the motion transfer system, which can transfer the user photo to the sign language video of specific words. In particular, the appearance content of the output video comes from the provided user image, while the motion of the video is extracted from the specified tutorial video. We observe two primary limitations in adopting the state-of-the-art motion transfer methods to sign language generation: 
(1) Existing motion transfer works ignore the prior geometrical knowledge of the human body. (2) The previous image animation methods only take image pairs as input in the training stage, which could not fully exploit the temporal information within videos. In an attempt to address the above-mentioned limitations, we propose Structure-aware Temporal Consistency Network  (STCNet) to jointly optimize the prior structure of human with the temporal consistency for sign language video generation. There are two main contributions in this paper. (1) We harness a fine-grained skeleton detector to provide prior knowledge of the body keypoints. In this way, we ensure the keypoint movement in a valid range and make the model become more explainable and robust. (2) We introduce two cycle-consistency losses, \ie, short-term cycle loss and long-term cycle loss, which are conducted to assure the continuity of the generated video. We optimize the two losses and keypoint detector network in an end-to-end manner. 
Empirical evaluation on the three widely-used datasets, \ie, LSA64, RWTH-PHOENIX-Weather 2014T, and WLASL-2000 verifies the effectiveness of the proposed method quantitatively and qualitatively. Besides, we also conduct extensive ablation studies to show that jointly learning the structure and temporal cues leads to high-fidelity video generation. We hope this work can pave the way for future studies on sign language generation, and social communication with fewer limitations.
\end{abstract}


\IEEEpeerreviewmaketitle

\section{Introduction}
\IEEEPARstart{A}{ccording} to \textit{Genesis 11:1-9}, humanity was stated to speak a single language but God smote people with the confusion of tongues due to the tower of Babel. Perhaps this is one of the old stories on the origin of different spoken languages. However, there is one  exception, \ie, sign language. Sign language is a typical type of visual language which conveys meanings via hand gestures and facial expressions \cite{tavella2022phonology}. Similar to traditional voice-based languages, sign language has broad usage scenarios including shopping, telecast, and education. People with hearing loss can express their emotions and opinions smoothly with the help of sign language \cite{cheok2019review}. According to the World Federation of the Deaf  (WFD), there are around 72 million people worldwide who usually use sign language \cite{murray2018education}. Moreover, almost all well-known TV news programs provide real-time sign language translation. Apparently, sign language is effective and influential around the world. 


Nevertheless, learning sign language could be time-consuming and difficult, which is beyond practicable for the public. That brings a gap between people with hearing loss and others. As with other traditional sound-based languages, sign language has a set of grammar and lexicon \cite{cheng2020fully}. However, sign language is affected by the local language and culture which means that sign language is not universal worldwide, making it harder for people from different regions to communicate fluently \cite{jiang2021sign}. To reduce the learning time for sign language and alleviate the misunderstanding situations caused by regional and cultural differences, we hope to animate a still image according to the sign language motion given by a tutorial video. The motion transfer technology could help people perform sign language without even learning.

\begin{figure}[tbp]
\centering
\animategraphics[autoplay,loop,width=0.95\linewidth]{10}{gif1/0}{010}{070}
\vspace{0.05in}
  \caption{\textbf{The dynamic show of sign language motion transfer results.} Given a source image and a driving video, the model generates a new video clip where the person in the source image performs the sign language motion in the driving video. Compared with the state-of-the-art method TPSMM \cite{zhao2022thin}, our method can generate smooth videos while preserving the identity attributes such as hair and face. {\color{citecolor} (Please open Adobe Reader to see the movement.)}
}
\label{fig:gif}
\vspace{-.2in}
\end{figure}

\begin{figure}[t]
\begin{center}
\includegraphics[width=1.0\linewidth]{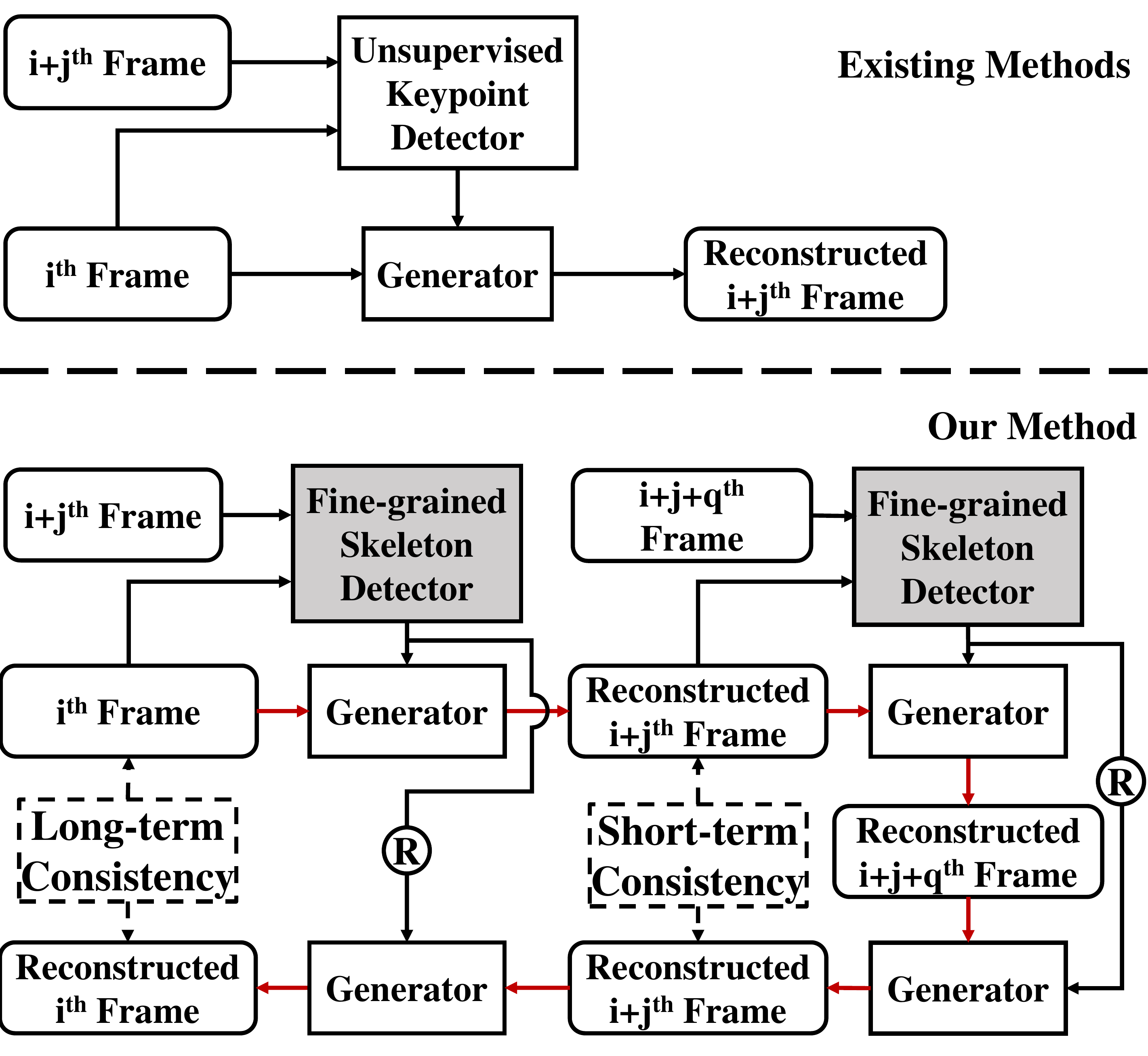}
\end{center}
   \caption{\textbf{Comparison between existing methods and our method.} Existing methods \cite{siarohin2019animating,siarohin2019first,siarohin2021motion,zhao2022thin} usually deploy an unsupervised keypoint detector and only perform a single frame generation procedure during training. 
   In contrast, our method explicitly considers both spatial and temporal information via harnessing the fine-grained skeleton detector and restricting two types of temporal consistency. 
   $\large{\textcircled{\small{R}}}$ in the figure indicates that we exchange the source image and the driving image to estimate the motion reversely. The \textcolor{red}{red arrows} show the generation order of our method during training.}
\label{fig:comparison}
\end{figure}

Over the past few years, significant progress has been made on motion transfer ~\cite{siarohin2019animating,siarohin2019first,siarohin2021motion,zhao2022thin}. Given a video and an image containing the same type of object, the goal of motion transfer is to generate a new video whose style comes from the image and the object motion comes from the video. 
We find that there are two main limitations when directly applying existing motion transfer methods to sign language generation. Firstly, the prior knowledge of body structure is underestimated. For instance, most works~\cite{siarohin2019animating,siarohin2019first,siarohin2021motion,zhao2022thin}  are usually based on unsupervised keypoint detection. Since the unsupervised keypoint detection is learned from scratch, the learned keypoints are usually not aligned with the semantic body parts. Especially, small-scale patterns, such as fingers, are usually missing or blurred (see Figure~\ref{fig:gif}). 
The second limitation is the lack of long-term temporal consistency. 
Existing works~\cite{siarohin2019animating,siarohin2019first,siarohin2021motion,zhao2022thin} encourage the short-term continuity between two frames.  
For instance, given a pair of images, this line of works focuses on the reconstructed single frame quality during training (see the top part of Figure~\ref{fig:comparison}) while ignoring more frames in the future. 

To address these limitations, we propose Structure-aware Temporal Consistency Network  (STCNet), a human body structure-aware network that generates sign language video smoothly.  
There are three features of the proposed framework. (1) We harness a fine-grained keypoint detector network to offer strong human body structure knowledge, enhancing hand motion estimation. (2) We propose the short-term cycle loss and the long-term cycle loss to promote the continuity of the generated videos. (3) Since the output of the keypoint detector network is unstable, we conduct a jointly training strategy to fine-tune the pre-trained keypoint detector network without extra annotations.
We conduct extensive experiments on an Argentinian sign language dataset named LSA64\cite{ronchetti2016lsa64}, a german sign language dataset called RWTH-PHOENIX-Weather 2014T \cite{camgoz2018neural}, and an American sign language dataset named WLASL-2000 \cite{li2020word}. 
The results suggest that our method surpasses state-of-the-art methods such as Monkey-Net \cite{siarohin2019animating}, First Order Motion Model  (FOMM) \cite{siarohin2019first}, Articulated Animation  (AA) \cite{siarohin2021motion}, and Thin-Plate Spline Motion Model  (TPSMM)\cite{zhao2022thin} with respect to the quality of the generated videos. Figure~\ref{fig:gif} shows that our method can generate smooth videos with correct motion details compared with the state-of-the-art method TPSMM.

Briefly, our contributions can be concluded as follows:
\begin{itemize}
    \item We identify practical problems that existing motion transfer methods
     (1) ignore the prior geometrical knowledge of human bodies and  (2) lack video continuity, especially in  a relatively long duration.   
     To address these two limitations, we propose a Structure-aware Temporal Consistency network  (STCNet). In particular, we explicitly introduce the prior human keypoints to guide the generation and involve the temporal consistency objective to further regularize the training process. 
    \item Extensive experiments on LSA64 \cite{ronchetti2016lsa64}, RWTH-PHOENIX-Weather 2014T\cite{camgoz2018neural}, and WLASL-2000 \cite{li2020word} datasets show that our approach surpasses several competitive methods, verifying the effectiveness of the proposed method qualitatively and quantitatively. The ablation studies also prove the effectiveness of jointly harnessing prior structure and temporal consistency for sign language video generation. 
\end{itemize}

The rest of the paper is organized as follows. Section~\ref{related} introduces existing related works, and the current research status is also discussed.  The proposed method, STCNet, is described in Section~\ref{Method}, and the network structure is illustrated in detail as well. Then, to validate our method, we conduct extensive experiments and the results are analyzed in Section~\ref{experiments} before we summarize our work in Section~\ref{conclusion}.

\section{Related Work} \label{related}

\subsection{Skeleton keypoint Detection}
Skeleton keypoint detection, also known as pose estimation, is to locate the essential parts of people in an image or a video \cite{dang2019deep}.  The intention of this work using skeleton keypoint detection is to guide the generation model via human body visual knowledge \cite{luo2018macro,yang2021multiple}. In particular, explainable keypoint locations help the generation model distinguish the occlusions caused by fingers and cloths. For example, the hand in front of the body is a kind of self-occlusion rather than the texture of the clothing.  
DeepPose is the pioneering work that applies deep learning in pose estimation \cite{toshev2014human} and outperforms traditional methods based on regression or retrieval \cite{dantone2013human,eichner20122d}. As for now, the state-of-the-art methods are generally derived from Convolutional Neural Networks \cite{cao2017realtime,papandreou2018personlab}.\par
In essence, all pose estimation methods can be divided into top-down methods and bottom-up methods \cite{zauss2021keypoint}. The bottom-up method means detecting joints first and gathering several joints to estimate the pose of a human. Representative work includes DeepCut \cite{pishchulin2016deepcut}, Associative Embedding \cite{newell2017associative}, PifPaf \cite{kreiss2019pifpaf}, OpenPifPaf \cite{kreiss2021openpifpaf}, Keypoint Communities \cite{zauss2021keypoint}, \emph{etc}. Deepcut \cite{pishchulin2016deepcut} is the first work to reformulate the pose estimation task into three problems: the selection of body parts, the labeling of the selected body parts, and the partitioning of the body parts that belong to the same person. Associative Embedding \cite{newell2017associative} improves the accuracy by predicting heatmaps and tag embeddings simultaneously. PifPaf, OpenPifPaf, and Keypoint Communities are a series of works where Keypoint Communities is the latest novel bottom-up method. Keypoint Communities \cite{zauss2021keypoint} models all the keypoint of a human as an overall graph consisting of many ego-centric graphs and uses community detection to compute the weights of every single keypoint. 
On the contrary, the top-down method means detecting a human first and then estimating the joints within the bounding box. CFN uses a "Coarse-Fine" network structure to exploit multi-level supervision for keypoint detection \cite{huang2017coarse}. CPN \cite{chen2018cascaded} introduces a cascaded pyramid network that aims to solve hard cases such as occluded keypoints. CrowdPose \cite{li_crowdpose_2019} designs a person-joint connection graph to deal with two main challenges in crowded scenes: wrong joint assembling and redundant pose prediction. 
RMPE, also known as Alphapose \cite{fang_rmpe_2017,li2020pastanet} designs Symmetric Spatial Transformer Network, Parametric Pose NonMaximum-Suppression, and Pose-Guided Proposals Generator to handle inaccurately detected bounding boxes. The merit of the top-down methods is the high accuracy and stability. However, the top-down methods are time-consuming compared to the bottom-up methods. The reason is that top-down methods have to do the crowd people detection first \cite{xiu2018poseflow}. Since every sign language video only contains one signer, we skip the human detection process and leverage the state-of-the-art AlphaPose method to extract key points.


\begin{figure*}[t]
\begin{center}
\includegraphics[width=1.0\linewidth]{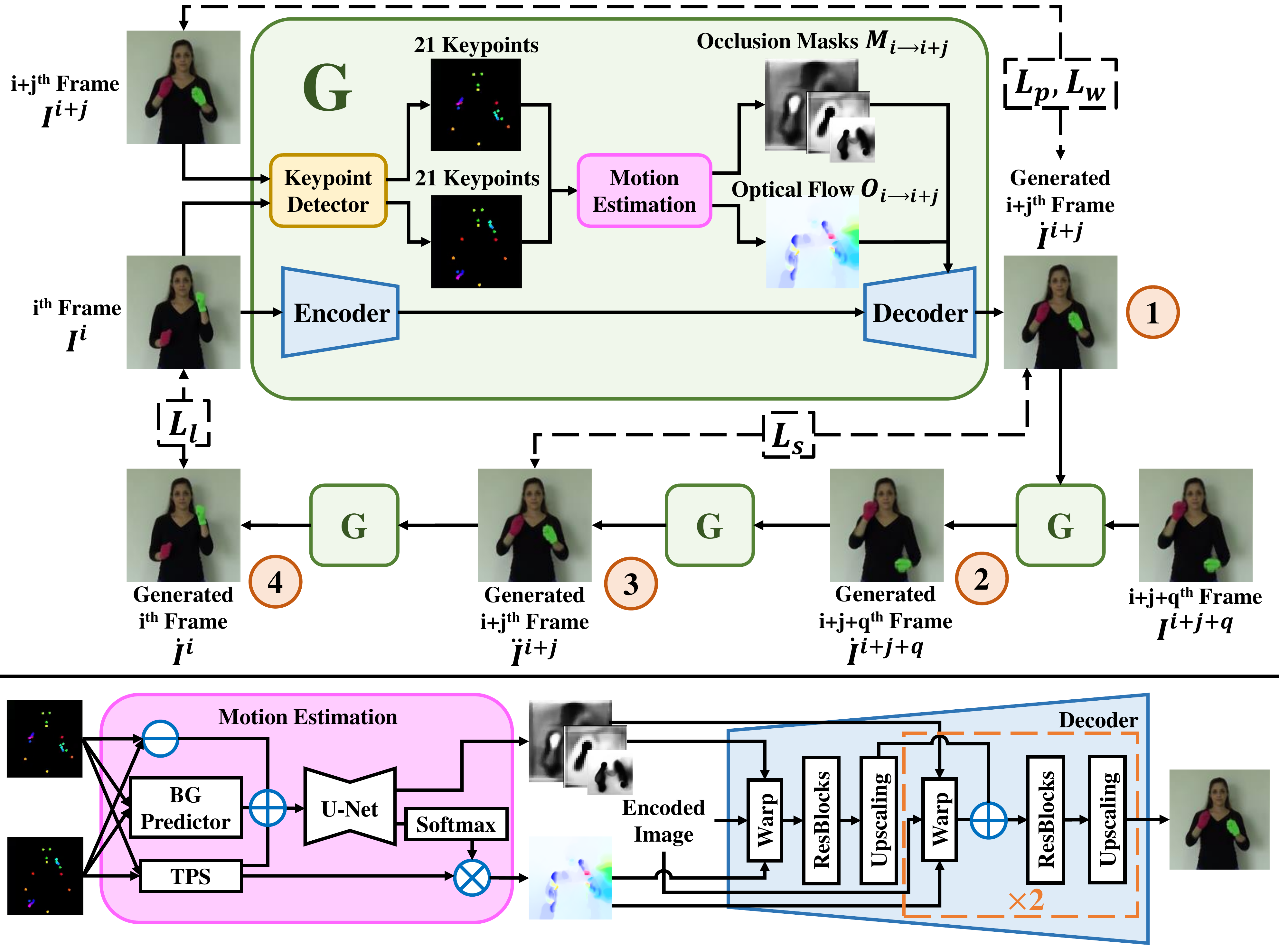}
\end{center}
   \caption{\textbf{The Brief Framework.} The generator  (G) consists of an encoder, a decoder, a keypoint detector, and a motion estimation network. Specifically, the keypoint detector takes the source image $I^i$ and the driving image $I^{i+j}$ as input and passes the location of the keypoints to the pose estimation network to predict the optical flow $O$ and the occlusion masks $M$. The encoder downsamples the source image to extract features whereas the decoder warps the encoded image according to the predicted optical flow $O$ and occlusion masks $M$ and generates the final output $\dot{I}^{i+j}$. For every iteration, the model takes three original frames in the same video clip as the input, \ie, the $i^{th}$ frame $I^i$, the $i+j^{th}$ frame $I^{i+j}$, and the $i+j+q^{th}$ frame $I^{i+j+q}$. As shown in the figure, we perform the generation procedure four times as a loop generation. The proposed losses and training strategy are also displayed in the figure. In particular, the short-term cycle loss and the long-term cycle loss are calculated between the generated frames. The detailed structure of the motion estimation network and the decoder is depicted at the bottom. Note that $\ominus$ here indicates element-wise subtract, $\oplus$ indicates concatenation, and $\otimes$ represents element-wise product.}
\label{fig:structure}
\end{figure*}

\subsection{Video Generation for Motion Transfer}
Video generation can be considered as generating a series of frames with spatial information. State-of-the-art video generation methods use neural networks as deep learning becomes trendy \cite{tulyakov2018mocogan,menapace2021playable}. GANs \cite{goodfellow2014generative} and VAEs \cite{DBLP:journals/corr/KingmaW13} lay the groundwork for future research\cite{9343766,9345477}. 
In this paper, we aim to transfer the motion in a sign language video to an image. Sign Language Production  (SLP) is a related research field which is to generate sign language videos  \cite{saunders2020everybody,saunders2020adversarial,saunders2020progressive,zelinka2020neural}.  However, this line of works studies the abstract text or audio inputs, which is different from the video-driven setting in this paper. 
Another related research field is motion transfer \cite{siarohin2019animating}. Motion transfer, also known as image animation, is to synthesize action given an image and a driving video and achieve "Do as I Do".  Efros \etal  \cite{efros2003recognizing} propose a retrieval-based action synthesis method that uses optical flow as a motion descriptor. In recent years, most motion transfer generation networks deploy the deeply-learned network\cite{9200521}. 
For instance, MoCoGAN \cite{tulyakov2018mocogan} adopts a motion and content decomposed representation for video generation by splitting the latent space into content and motion subspaces so that videos in different motions and contents can be generated by sampling different points in the different subspaces. 
Yang \etal propose a two-stage approach, \ie, PSGAN and SCGAN \cite{yang2018pose}, to transfer motions collaboratively. PSGAN first yields pose sequences, then SCGAN generates video frames from the poses. 
Everybody Dance Now  (EDN) \cite{chan2019everybody} is another two-stage motion transfer network. The first stage is to generate the image frame as a whole and the second stage is to add more detail and realism to the face region. Zhou \etal  propose another two-stage dance motion transfer network using a spatial transformer network \cite{zhou2019dance}. \textsc{G$^{3}$AN} \cite{wang2020g3an} is a three-stream video generation network that learns to disentangle appearance features, motion features, and smoothing generated videos. Siarohin \etal  propose a series of works on motion transfer including Monkey-Net \cite{siarohin2019animating}, First Order Motion Model  (FOMM) \cite{siarohin2019first}, and Articulated Animation (AA) \cite{siarohin2021motion}. Monkey-Net is an end-to-end motion transfer network, which learns to detect keypoint in an unsupervised way. FOMM is an extension of Monkey-Net that calculates the first-order Taylor expansion in a neighborhood of the keypoint locations. Instead of using existing works on optical flow estimation \cite{6884847}, Monkey-Net and FOMM predict the optical flow based on the location of the keypoints. Unlike the keypoint-based methods, AA transfers the motion from the essential regions of the object. Recently, Zhao \etal  propose a framework that focuses on human face animation \cite{Zhao_2021_ICCV}. Liu \etal  adopt neural-ODEs for motion deformation \cite{liu2021self}. Yoon \etal  design a network to animate images using UV maps produced by a 3D human model \cite{yoon2021pose}. For the appearance transfer, some early works, such as MUNIT~\cite{huang2018multimodal} and DG-Net~\cite{zheng2019joint}, apply the Adaptive Instance Normalization layer (AdaIN)~\cite{huang2017arbitrary} to enable the color changing, but it is not good at changing the pattern location largely or keeping fine-grained patterns, like logo~\cite{zheng2019joint}. 
In contrast, Thin-Plate Spline Motion Model  (TPSMM) by Zhao \etal \cite{zhao2022thin} applies Thin-Plate Spline  (TPS) transformation based on FOMM, which preserves the fine-grained patterns such as hands. However, TPSMM needs five sets of keypoint which usually bring redundancy.
In contrast, our work is different from existing works in several aspects:  (1) We explicitly introduce the keypoint detection network with the geometrical meaning of human upper part body structure, and further harness such points to provide the prior guidance of generation.
 (2) Inspired by the spirit of the cycle consistency~\cite{wang2019learning}, we consider the temporal consistency could be an essential influence factor of the generated video quality.  (3) We jointly optimize the spatial and temporal consistency, yielding high-fidelity sign language video generation.

\section{Method} \label{Method}

We propose STCNet, a body structure-aware framework that focuses on sign language motion transfer while maintaining the cycle consistency of time. In this section, we illustrate the network structure and introduce the training strategy in detail. We depict the difference between our method and existing methods in Figure~\ref{fig:comparison}, then show the framework structure in Figure~\ref{fig:structure}. STCNet consists of four parts, a keypoint detector network  (Section~\ref{KDM}), a motion estimation network  (Section~\ref{MGM}), an encoder, and a decoder  (Section~\ref{ED}). The keypoint detector provides body structure knowledge by detecting 21 landmarks on the upper part body of the signer. The motion estimation network takes the keypoint locations as input and predicts the optical flow and the occlusion masks. The encoder extracts feature maps in different resolutions whereas the decoder warps the feature map according to the optical flow and generates a new image simulating the objective frame. After introducing the network structure, we discuss the optimization goals and the training testing strategy in Section~\ref{OP}.

\subsection{Keypoint Detector Network}\label{KDM}
To obtain explainable and accurate keypoint locations, we fine-tune an  Alphapose model pre-trained on the Halpe dataset \cite{li2020pastanet}. Halpe dataset is a large whole-body dataset with 136 annotated landmarks on the human body. There are 68 keypoints on the face, 26 on the body, and 42 on the hands. Such a fine-grained keypoint annotation offers strong supervision, helping the detection model captures subtle facial expressions and hand motions. Sign language videos only contain the upper part human body, and the semantic information is revealed by the hand gestures and the facial expressions of the signer. Therefore, we select 21 vital keypoints with 12 on the hands, 5 on the upper body, and 4 on the face to remove redundancy without losing the details. 
Given the $i^{th}$ frame $I^i \in \mathbb{R}^{H \times W \times 3}$ in a video clip, we pass the image to the Alphapose model $F_P$ and get the result coordinates $K^i \in \mathbb{R}^{21 \times 2}$. The keypoint detection procedure can be formulated as follow:
\begin{equation}
  \label{eq:3}
    K^i = F_P (I^i). 
\end{equation}
Empirically, we find that the output of two continuous frames within a video clip generated by the pre-trained AlphaPose model differs a lot, especially when the frames are vague. One possible reason behind the large discrepancy between the detection results could be lacking continuity of time. To address this problem, we fine-tune the keypoint detector along with the training procedure of the other modules. The keypoint detector shares the optimization goals with the other modules, thus no extra annotation is required. A violent fine-tuning procedure leads to missing the body structure information provided by the pre-trained model. Hence, we set the learning rate at a  smaller value to fine-tune the detection module slowly. 

\subsection{Motion Estimation Network}\label{MGM}
We compare multiple works on the image warping method using keypoint \cite{siarohin2019animating,siarohin2019first,Zhao_2021_ICCV,liu2021self,zhao2022thin}. Considering the effectiveness and efficiency, we deploy a network based on Thin Plate Spline (TPS) transformation. The motion estimation network aims to predict a dense optical flow $O_{i \rightarrow i+j} \in \mathbb{R}^{H/4 \times W/4 \times 2}$ indicating the motion of the upper part of the body. $i \rightarrow i+j$ means the model takes the $i^{th}$ frame $I^i$ as the source image and the $i+j^{th}$ frame $I^{i+j}$ as the driving image.

Specifically, given 21 pairs of keypoint detected from the source image and driving image, we first use Thin Plate Spline (TPS) transformation \cite{bookstein1989principal} to estimate 21 corresponding optical flows. 
A learnable background predictor is adopted to predict an extra optical flow to approximate the motion of the background \cite{siarohin2021motion, zhao2022thin}. 
We warp the downsampled source image according to each coarse optical flow mentioned above for later use.
Every keypoint is modeled by a Gaussian in a heatmap, which means two sets of heatmaps can be obtained from the source image and the driving image. 
To emphasize the keypoint location difference between the source image and the driving image, the previously warped images are concatenated with the heatmap difference and then used as the input of the U-net structure \cite{ronneberger2015u} to learn the residual motions. The output of the U-Net network is passed to a softmax layer and then multiplied with the coarse optical flows elementwisely. The final predicted optical flow $O_{i \rightarrow i+j}$ is obtained by summing the multiply result along the channel axis.
Meanwhile, the motion estimation network additionally predicts a set of occlusion masks $M$ in different resolutions via applying a convolutional layer after the U-Net structure. The occlusion masks are applied in the decoder network to mask the unnecessary deformation within the feature map. Overall, the motion estimation process can be summarized as: 
\begin{equation}
\label{eq:3}
\begin{aligned}
    &M_{i \rightarrow i+j} = F_M (F_U (K^{i}, K^{i+j})), \\
    &O_{i \rightarrow i+j} = F_O (F_U (K^{i}, K^{i+j})), 
\end{aligned}
\end{equation}
where $F_U$ denotes the image representation processing function. In practice, we leverage a U-Net \cite{ronneberger2015u} structure as $F_U$ to preserve the spatial patterns. $F_M$ is a $7\times7$ convolutional layer used to predict the occlusion masks $M_{i \rightarrow i+j}$ in different resolutions, and $F_O$ represents a softmax layer followed by multiplying the optical flows estimated by TPS, and a sum operation along the channel axis. 
Both $F_M$ and $F_O$ take the output of the U-Net model as the input.
\par
\subsection{Encoder and Decoder}\label{ED}
\noindent\textbf{Encoder.} As Figure~\ref{fig:structure} shows, the source image $I^i \in \mathbb{R}^{H \times W \times 3}$ is first passed to the encoder to extract features. We adopt a simple but effective "high to low" architecture for the encoder. The intuition is to combine the general information in the low-resolution feature maps and the detailed information in the high-resolution feature maps. The input image $I^i$ is first passed to a convolutional layer to expand the feature channel. Followed by three downsampling blocks, the encoder aims to capture high-level features step by step. Every downsampling block consists of a $3\times3$ convolutional layer with a stride of 1, an instance normalization layer, a ReLU activation layer, and a $2\times2$ average pooling layer with a stride of 2.  

\noindent\textbf{Decoder.} Let $F_E$ denote the encoder model and $F_D$ denote the decoder model. To get the generation result $\dot{I}^{i+j} \in \mathbb{R}^{H \times W \times 3}$, the deformation and decoding processes are carried on concurrently and progressively and can be formulated as:
\begin{equation}
  \label{eq:3}
    \dot{I}^{i+j} = F_D (F_E (I^{i}), M_{i \rightarrow i+j}, O_{i \rightarrow i+j}). 
\end{equation} 
As shown in the bottom of the Figure~\ref{fig:structure}, the encoded feature map in the lowest resolution (\ie, the output of the third downsampling block in the encoder) is first warped according to the optical flow $O_{i \rightarrow i+j}$ and then multiplied by the occlusion mask $M_{i \rightarrow i+j}$ in the corresponding resolution. Followed by a series of Resblocks, the model learns the residual information. In particular, a Resblock contains two $3 \times 3$ convolutional layers and a shortcut connection \cite{he2016deep}. The input of each convolutional layer in the Resblock is normalized by an instance normalization layer and activated by a ReLU layer. We then warp the middle-resolution feature map (\ie, the output of the second downsampling block in the encoder) according to the optical flow $O_{i \rightarrow i+j}$ and multiply the result by the occlusion mask. The warped middle-resolution feature map is concatenated with the upsampled output of the previous Resblocks and then passed to another series of Resblocks. Likewise, the feature map in the highest resolution  (\ie, the output of the first downsampling block in the encoder) is also warped by the optical flow, multiplied by the occlusion mask, concatenated with the output of the previous Resblocks, and passed to some other Resblocks. The decoding process ends up with a final convolutional layer which decreases the channel number to three. We use a sigmoid layer to restrict the output value and get the final result $\dot{I}^{i+j}$. Similarly, we can get the result of the other generation procedures depicted in Figure~\ref{fig:structure}, which can be formulated as: 
\begin{equation}
\label{eq:4}
\begin{aligned}
    &\dot{I}^{i+j+q} = F_D (F_E (\dot{I}^{i+j}), M_{i+j \rightarrow i+j+q}, O_{i+j \rightarrow i+j+q}),\\
    &\ddot{I}^{i+j} = F_D (F_E (\dot{I}^{i+j+q}), M_{i+j+q \rightarrow i+j}, O_{i+j+q \rightarrow i+j}),\\
    &\dot{I}^{i} = F_D (F_E (\ddot{I}^{i}), M_{i+j \rightarrow i}, O_{i+j \rightarrow i}).  
\end{aligned}
\end{equation}

\begin{figure*}[t]
\begin{center}
\includegraphics[width=1.0\linewidth]{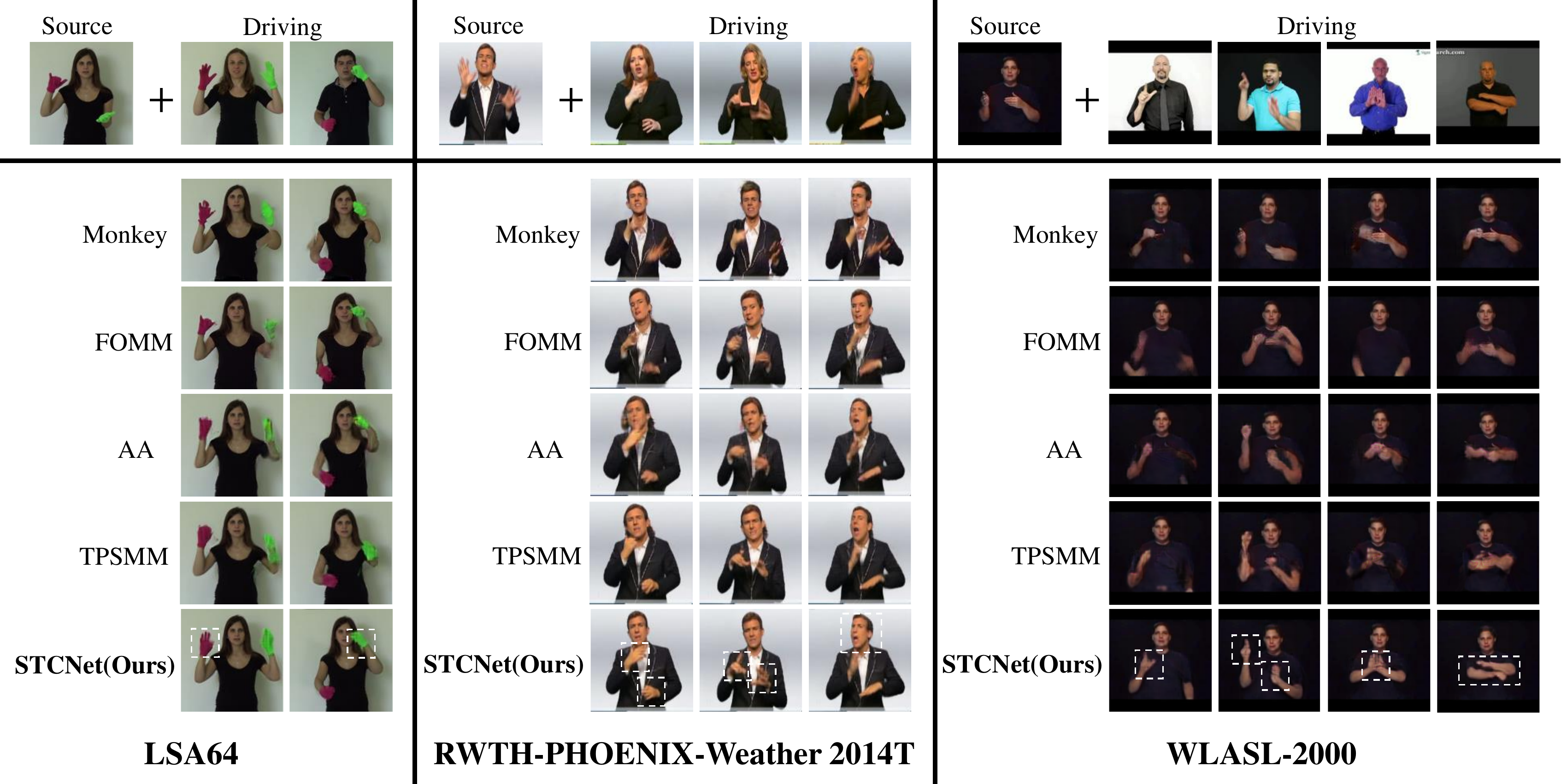}
\end{center}
   \caption{\textbf{Qualitative comparison.} We compare our method with previous methods under the transfer setting on the three datasets, \ie, LSA64~\cite{ronchetti2016lsa64}, RWTH-PHOENIX-Weather2014T~\cite{camgoz2018neural}, and WLASL-2000~\cite{li2020word}. We present the visual result of the same source image driven by different driving images. Note that the identity of the driving image is different from the identity of the source image. Our method preserves the identity attributes of the source image and provides fine-grained motion details extracted from the driving image (highlighted in the white dashed boxes).}
\label{fig:cherry}
\end{figure*}

\subsection{Optimization}\label{OP}
During training, we apply a compound reconstruction loss as the optimization goal, which is an aggregation of widely used losses to supervise the model to generate output with the same gesture and facial expression as the driving frame. Pyramid perceptual loss, middle feature loss, short-term cycle loss, and long-term cycle loss are the main components of the reconstruction loss. It is worth mentioning that the keypoint detector network is jointly optimized by the same loss. Moreover, we also introduce the training strategy and the reconstruction policy in detail.

\noindent\textbf{Pyramid Perceptual Loss.}
Perceptual loss is proposed by Johnson~\etal and is widely used in image transformation and reconstruction \cite{johnson2016perceptual}. The main idea is to supervise the generation network using the feature maps extracted by a pre-trained image classification network. In our work, a pre-trained VGG-19 network is deployed to extract high-quality features on multi-resolution input images. We assume that similar images should have similar feature maps. Therefore we minimize the L1 distance between two feature maps in five different middle layers, which offers stronger supervision than traditional image-level reconstruction losses such as MSE loss. The loss can be depicted as:
\begin{equation}
\mathcal{L}_p=\sum_{n}\left|V^{n} (I^{i+j})-V^{n} (\dot{I}^{i+j})\right|,
\end{equation}
where $I^{i+j}$ denotes the ground truth driving image and $\dot{I}^{i+j}$ means the generated image. $V^n$ represents the $n^{th}$ layer output of the pre-trained VGG-19 network~\cite{vgg}.
In practice, we downsample the image pair and conduct a pyramid perceptual loss to further facilitate the reconstruction supervision in different resolutions \cite{siarohin2019animating}. \par
\noindent\textbf{Warp Consistency Loss.}
Apart from the supervision of the generation results, we further constrain the warped encoded image to simulate the encoded driving image in the generation network \cite{zhao2022thin}. To this end, warp consistency loss is defined as:
\begin{equation}
\mathcal{L}_w=\sum_{r}\left|\mathcal{W} (F_E^r (I^i), O_{i \rightarrow i+j})-F_E^r (I^{i+j})\right|.
\end{equation}
Note that $I^i$ denotes the source image, and $I^{i+j}$ represents the driving image. $F_E^r$ means the $r^{th}$ downsampling block in the encoder architecture. $\mathcal{W}$ is the warping operation according to the predicted optical flow $O_{i \rightarrow i+j}$. It is worth mentioning that the warping operation uses a bilinear sampler and does not contain any learnable parameters.
\par
\noindent\textbf{Cycle-Consistency Losses.}
Temporal continuity is an essential influence factor for video generation since the real world is smooth and coherent. The previous two loss functions provide supervision on the image level, which ignores the temporal continuity in a video clip. To address this problem, we propose two types of cycle-consistency losses: short-term cycle loss and long-term cycle loss. The short-term cycle loss is defined as:
\begin{equation}
\mathcal{L}_s=\left|\dot{I}^{i+j}-\ddot{I}^{i+j}\right|.
\end{equation}
As Figure~\ref{fig:structure} presents, $\dot{I}^{i+j}$ is the image generated by the first generation procedure, which means the model takes the $i^{th}$ frame $I^i$ as the source image and the $i+j^{th}$ frame $I^{i+j}$ as the driving image. $\ddot{I}^{i+j}$ is the image generated by the third generation procedure. Although sharing the same driving image with the first generation procedure, the third procedure considers the output of the second generation procedure $\dot{I}^{i+j+q}$ as the source image. Based on the temporal consistency hypothesis, the short-term cycle loss minimizes the $L_1$ distance between $\dot{I}^{i+j}$ and $\ddot{I}^{i+j}$. In other words, the short-term cycle loss grants the model the ability to generate the same results given different source images and the same driving image. To further promote the cycle-consistency of time, we adopt the long-term cycle loss which is an augmented version of the original cycle loss. The long-term cycle loss is defined as:
\begin{equation}
  \label{eq:7}
    \mathcal{L}_l=\left|\dot{I}^i - I^i\right|.
\end{equation}
The long-term cycle loss performs a larger cycle compared to the previous short-term cycle loss as shown in Figure~\ref{fig:structure}. We aim to consolidate the cycle consistency by minimizing the $L_1$ distance between the output of the fourth generation procedure $\dot{I}^i$ and the $i^{th}$ frame $I^i$. Overall, the short-term cycle loss and the long-term cycle loss provide temporal self-supervision, promoting the continuity and consistency of the generated videos.
\par
The overall loss function is a combination of the above losses:
\begin{equation}
\mathcal{L}_{total}=\lambda_p\mathcal{L}_p + \lambda_w\mathcal{L}_w + \lambda_c (\mathcal{L}_s + \mathcal{L}_l).
\end{equation}
The pyramid perceptual loss is considered the most essential reconstruction loss, thus we follow existing works~\cite{zhao2022thin} and set the weight $\lambda_p$ to 10. $\lambda_w$ represents the weight of the warp consistency loss and is set to 1. The short-term cycle loss and long-term cycle loss share the same weight. We conduct a grid search of  0.5, 1, 2, 5, 10 for $\lambda_c$ and 2 turns out to be the optimal weight for the cycle-consistency losses. It is because too high cycle-consistency losses lead to a trivial solution that all generated images are not changed as the input $I^i$. Having this hybrid optimization goal, the generation network can generate high-fidelity videos while maintaining the cycle consistency of time.

\noindent\textbf{Training Strategy.} As Figure~\ref{fig:comparison} shows, existing motion transfer networks usually take an image pair as input and perform the generation procedure once to calculate the reconstruction loss. This training mode neglects the temporal information in a video clip, thus the generation results vary when given the same driving image but different source images picked from the same video. To address this problem, our framework conducts a cyclic end-to-end jointly training strategy as shown in Figure~\ref{fig:structure}. For every iteration, the model randomly chooses three frames from a video clip as input and performs the generation procedure four times. The short-term cycle loss is calculated between $\dot{I}^{i+j}$ and $\ddot{I}^{i+j}$, while the long-term cycle loss is calculated between $\dot{I}^i$ and $I^i$. We consider the temporal cycle-consistency as the prior hypothesis which helps the model learn the temporal information. The benefit of our training strategy is that the models can promise strong temporal robustness and generate videos with high continuity. Every rose has its thorn. The proposed strategy could be time-consuming in one iteration but converges quickly overall. \textbf{Since the pre-trained image-based keypoint detector network does not maintain video continuity when facing blurred frames in a video clip, we jointly optimize the keypoint detector network and the generator network using the same optimization objective as mentioned earlier.} The short-term cycle loss and the long-term cycle loss empower the model with cycle consistency of time without extra human body structure annotations, making the joint training strategy optimal and data-efficient. Nevertheless, overfitting to the cycle-consistency losses makes the keypoint detector lose its original pre-trained geometric prior knowledge. To balance the consistency penalty and the body structure knowledge, we set the learning rate of the keypoint detector network to a small value.

\noindent\textbf{Inference Strategy.} 
Similar to the training stage, the model takes two images as input for every generation procedure during testing. The model generates a new image, which resembles the target motion by deforming the source image. To generate the whole video clip, we further use the first image of a video clip as the source image and other frames as the driving images in sequence. Some metrics are applied to evaluate the generation quality of the generated images. Except for this reconstruction scenario, the testing method can also be extended to the situation where the source image and the driving image are not from the same video. Our model can extract the motion information from the driving image and transfer the motion to the source image. Extensive qualitative and quantitative results are shown in the next chapter.

\begin{table*}[t]
\caption{Results for the four competitive methods and our method under the reconstruction setting on the three datasets. Three image quality evaluation metrics are adopted to test the reconstruction ability of  models. Generally, $L_1$ is for the pixel level reconstruction, SSIM focuses on the local patterns, while LPIPS reflects the inception score by a deeply-learned model. We train the same epochs on every competitive method and the results show that the proposed method outperforms existing methods. }
\setlength{\tabcolsep}{2.8pt}
\begin{center}
\setlength{\tabcolsep}{4.0mm} 
\renewcommand{\arraystretch}{1.3}{
\begin{tabular}{l|c|c|c|c|c|c|c|c|c}
\shline
\multirow{2}{*}{Methods}& 
\multicolumn{3}{c|}{LSA64~\cite{ronchetti2016lsa64}} &
\multicolumn{3}{c|}{RWTH-PHOENIX-Weather 2014T~\cite{camgoz2018neural}} &
\multicolumn{3}{c}{WLASL-2000~\cite{li2020word}} \\
\cline{2-10}
& $L_1 \downarrow$ & SSIM $\uparrow$ & LPIPS $\downarrow$ & $L_1 \downarrow$ & SSIM $\uparrow$ & LPIPS $\downarrow$ & $L_1 \downarrow$ & SSIM $\uparrow$ & LPIPS $\downarrow$ \\
\shline

Monkey-Net~\cite{siarohin2019animating} & 0.0121 & 0.9489 & 0.0217 & 0.0340 & 0.8314 & 0.0784 & 0.0242 & 0.8786 & 0.0623\\
FOMM~\cite{siarohin2019first} & 0.0186 & 0.9151 & 0.0274 & 0.0253 & 0.8681 & 0.0461 & 0.0260 & 0.8726 & 0.0587\\
AA~\cite{siarohin2021motion} & 0.0110 & 0.9493 & 0.0187 & 0.0190 & 0.9077 & 0.0365 & 0.0178 & 0.9118 & 0.0415\\
TPSMM~\cite{zhao2022thin} & 0.0109 & 0.9499 & 0.0203 & 0.0188 & 0.9149 & 0.0327 & 0.0158 & 0.9218 & 0.0374\\

\hline
Ours & \textbf{0.0104} & \textbf{0.9533} & \textbf{0.0170} & \textbf{0.0172} & \textbf{0.9211} & \textbf{0.0302} & \textbf{0.0153} & \textbf{0.9273} & \textbf{0.0313}\\

\shline
\end{tabular}
}
\end{center}

\label{table:reconstruction}
\end{table*}
\section{Experiments} \label{experiments}
\subsection{Data Preparation}
\noindent\textbf{LSA64} \cite{ronchetti2016lsa64} is a small-scale dataset for the sign language recognition task containing 64 words in Argentinian Sign Language  (LSA). LSA64 consists of 3200 sign language videos performed by 10 different signers. Every signer performs each word 5 times. Interestingly, the signers wear red gloves on their right hands and green gloves on their left hands. The original resolution of the frames is $1920 \times 1080$, which is too large for our computing resources. Therefore, we crop the image to reduce the redundant background and resize the frames to $128 \times 128$. Moreover, we cut off the beginning and the end of every video since there is no movement. We randomly select 8 word categories from 64 classes as the test set and the remaining 56 categories as the training set. Therefore, there are 400 test videos and 2800 training videos. 
\par
\noindent\textbf{RWTH-PHOENIX-Weather 2014T} \cite{camgoz2018neural} is a German sign language dataset with 1066 different signs for the sign language translation problem. 
All the videos come from the weather forecasts editions. The dataset consists of 7738 videos performed by 9 different signers wearing dark clothes in front of a grey background. The original frame size is $210 \times 256$ and we resize the frame size to $128 \times 128$. In terms of the train-test split, we follow the setting in the original dataset where 7096 videos are used for training and the rest 642 videos for testing.
\par
\noindent\textbf{WLASL-2000} \cite{li2020word} is a large-scale word-level American Sign Language  (ASL) dataset including around 2000 words performed by more than 100 signers. WLASL-2000 is the largest publicly available ASL dataset so far. Different from LSA64 and RWTH-PHOENIX-Weather 2014T, videos in WLASL-2000 have different backgrounds which increases the difficulty of inference. The videos are collected from 20 different websites, promoting the diversity of data sources. There are 21083 videos in total and we use the official train-test split. The resolution of frames is resized to $128 \times 128$ as well.
\par

\subsection{Evaluation Metrics}
\noindent\textbf{Manhattan Distance ($L_1$)} \cite{vadivel2003performance} is the mean $L_1$ distance between every pixel of the generated frame and the ground-truth frame. 
It indicates the reconstruction ability of the trained model.
A lower $L_1$ distance indicates higher reconstruction quality. 
\par
\noindent\textbf{Structural SIMilarity  (SSIM)}\cite{2004Image} is a widely used metric for image quality evaluation. 
SSIM compares the resemblance between two images concerning luminance, contrast, and structure. The value of SSIM ranges from -1 to 1, where 1 means the two images are exactly the same. We calculate SSIM between the generated frame and the ground-truth frame, thus a higher value means higher generation quality.
\par
\noindent\textbf{Learned Perceptual Image Patch Similarity  (LPIPS)}\cite{zhang2018perceptual} proposed by Zhang \etal  is a metric used to compare the perceptual similarity between two images. We adopt the default Alexnet \cite{krizhevsky2012imagenet} version here. Note that a lower value of LPIPS means a higher similarity between the generated frame and the ground-truth frame.
\par
\subsection{Implementation Details}
We deploy a single Tesla V100 GPU to train models on every dataset. 
According to the dataset size, we train 100 epochs, 200 epochs, and 300 epochs for LSA64, RWTH-PHOENIX-Weather 2014T, and WLASL-2000 respectively. 
Therefore, the training process lasts for about one GPU day, two GPU days, and three GPU days correspondingly. Following existing works~\cite{zhao2022thin,siarohin2021motion}, the generator network and the motion estimation network are trained by the Adam Optimizer~\cite{diederik2014adam} with $\beta_1 = 0.5, \beta_2 = 0.99$. We set the initial learning rate to 0.0002 except for the keypoint detector network and apply a multistep scheduler to decay the learning rate. The learning rate decay factor $\gamma$ is set to 0.1. The decay happens at epochs 70 and 90 when training the LSA64 dataset, epochs 140 and 180 when training the RWTH-PHOENIX-Weather 2014T dataset, and epochs 210 and 270 when training the WLASL-2000 dataset. 
To balance the memory cost and the training speed, we adopt a batch size of 16. In terms of the keypoint detector network, we set the initial learning rate to 0.00002 while the decay happens along with the generator network. Since there is only a single human in the center of every frame of the dataset, we remove the human detection module from the off-the-shelf Alphapose pipeline \cite{fang_rmpe_2017,li2020pastanet} to boost the keypoint detecting speed and reduce training costs. The parameter number of the final model is about 104.83 M.

\subsection{Quantitative Results}
We compare our method with four previous representative works for motion transfer, including Monkey-Net \cite{siarohin2019animating}, FOMM \cite{siarohin2019first}, AA  \cite{siarohin2021motion}, and TPSMM \cite{zhao2022thin}. One line of works, including Monkey-Net, FOMM, and TPSMM is the keypoint-based method, which is similar to our method. The other line of works, such as AA, is the region-based approach that estimates human joint regions in an unsupervised manner. 
We adopt the best setting of these competitive methods reported in their papers for comparison. 
We evaluate the video reconstruction ability of all methods by three metrics mentioned earlier which are $L_1$, SSIM, and LPIPS. As shown in Table~\ref{table:reconstruction}, the proposed method surpasses other methods and reaches the state-of-the-art results of all three metrics on three datasets. 
The results verify that our method can generate video in high fidelity by recovering more human body structure details. 
The reason is that our keypoint detector learns 21 explainable keypoints, making the motion estimation model focus on the motion of essential body parts accurately. 
For instance, our method predicts 12 keypoints located on the hands, which explicitly makes the motion estimation network pay attention to fine-grained finger motions.

\subsection{Qualitative Results}
Different from the reconstruction setting in the quantitative comparison, we test the transferability of our method via qualitative analysis. As Figure~\ref{fig:cherry} shows, we select a source image and multiple driving videos with different identities to compare the animation quality (details highlighted in the white dashed boxes). 
For all three datasets, our method maintains high identity consistency and correct motion, especially in facial expression and hand details. Monkey-Net \cite{siarohin2019animating} and FOMM \cite{siarohin2019first} show poor motion transfer capability on every dataset. The body structure is even disconnected in some situations. AA \cite{siarohin2021motion} and TPSMM \cite{zhao2022thin} have a robust capability to capture the motion and preserve the correct body structure. However, the identity attributes of the generated image, such as hair and face, are relatively blurred with the identity of the driving image, especially on the RWTH-PHOENIX-Weather 2014T dataset. 
Our method surpasses these methods on the single frame quality especially with respect to hand motion details. 
It is because the pre-trained keypoint detector model explicitly raises the attention of the motion estimation network toward the hand motions and the facial expressions, providing more details in the generation results. 

Apart from the single-frame fidelity, we also provide visual results for the video continuity comparison. Since our method considers both the long-term and short-term cycle consistency, the motion in the reconstructed video is smoother than in the videos generated by other methods. We display some consecutive frames in Figure~\ref{fig:gif}. The reason behind the continuity is that we conduct the end-to-end training strategy which empowers the generator network with strong temporal consistency. 

\subsection{Ablation Studies}

\textbf{Do the proposed cycle-consistency losses work?} 
Table~\ref{table:ablation} shows the results of adopting the short-term cycle loss and long-term cycle loss on the LSA64 dataset. 
In particular, the $L_1$, SSIM, and LPIPS of the vanilla model trained without the two losses arrive at 0.0106, 0.9516, and 0.0179 respectively. We could observe two points: (1) Compared with the vanilla model, the short-term cycle loss and the long-term cycle loss can individually boost the reconstruction quality of the trained model. The short-term cycle loss has a larger regularization impact on the reconstruction.
(2) The short-term and long-term consistency losses are complementary. The model achieves the best performance (-0.0002 for L1, +0.0017 for SSIM, and -0.0009 for LPIPS) when both the two proposed losses are deployed. 
We think both losses ensure video continuity and robustness by refraining from the unrecoverable movements. 
\par

\setlength{\tabcolsep}{5pt}
\begin{table}[t]{
\caption{Ablation study on the proposed short-term cycle loss and the long-term cycle loss. We train the models while removing the losses in turn on the LSA64 dataset and then test the reconstruction ability using three metrics. Results verify the effectiveness of the proposed cycle-consistency losses. } 
\label{table:ablation}
{

\setlength{\tabcolsep}{5.5mm} 
\renewcommand{\arraystretch}{1.2}{
\begin{tabular}{cc|c|c|c}
\shline
$L_l$ & $L_s$ & $L_1 \downarrow$ & SSIM $\uparrow$ & LPIPS $\downarrow$ \\
\shline
 & & 0.0106 & 0.9516 & 0.0179 \\
 $\checkmark$ & & 0.0105 & 0.9519 & 0.0177 \\
 & $\checkmark$ & 0.0105 & 0.9526 & 0.0175 \\
$\checkmark$ & $\checkmark$ & 0.0104 & 0.9533 & 0.0170 \\
\shline
\end{tabular}}
}}
\end{table}

\textbf{Is the model sensitive to the cycle-consistency losses?} 
To test whether the model is sensitive to the weight of the cycle-consistency losses, we apply another experiment to explore the proper shared weight $\lambda_c$. As shown in Table~\ref{table:weight}, we attempt five different values of the weight including 0.5, 1, 2, 5, and 10. Under the same reconstruction setting carried out in the quantitative section, 2 is the optimal value for $\lambda_c$. We find that the model is not sensitive to the value of $\lambda_c$, yet a too large or too small value of $\lambda_c$ could harm the performance. For instance, compared with setting $\lambda_c$ to 2, the performance decreases when setting $\lambda_c$ to 0.5 (+0.0002 $L_1$, -0.0020 SSIM and +0.0009 LPIPS) or 10 (+0.0001 $L_1$, -0.0015 SSIM and +0.0007 LPIPS). We consider the reason is that lower weight for the cycle losses does not offer enough penalty on the cycle consistency. Meanwhile, a way larger weight could force the model to overfit the cycle losses and weaken the supervision provided by the reconstruction loss. Hence, we select 2 as the value of $\lambda_c$ to balance the influence of the cycle losses and the reconstruction loss.

\setlength{\tabcolsep}{5pt}
\begin{table}[t]{
\caption{Ablation study on the weight of the proposed losses. We train the same model using different $\lambda_c$ values and test the reconstruction ability on the LSA64 dataset. } 
\label{table:weight}
{

\setlength{\tabcolsep}{7.2mm} 
\renewcommand{\arraystretch}{1.2}{
\begin{tabular}{c|c|c|c}
\shline
$\lambda_c$ & $L_1 \downarrow$ & SSIM $\uparrow$ & LPIPS $\downarrow$ \\
\shline
0.5  & 0.0106 & 0.9513 & 0.0179 \\
1  & 0.0105 & 0.9528 & 0.0176 \\

2 & \textbf{0.0104} & \textbf{0.9533} & \textbf{0.0170} \\
5 & 0.0106 & 0.9518 & 0.0175 \\

10 & 0.0105 & 0.9518 & 0.0177 \\
\shline
\end{tabular}}
}}
\end{table}

\begin{figure}[t]
\begin{center}
\includegraphics[width=1.0\linewidth]{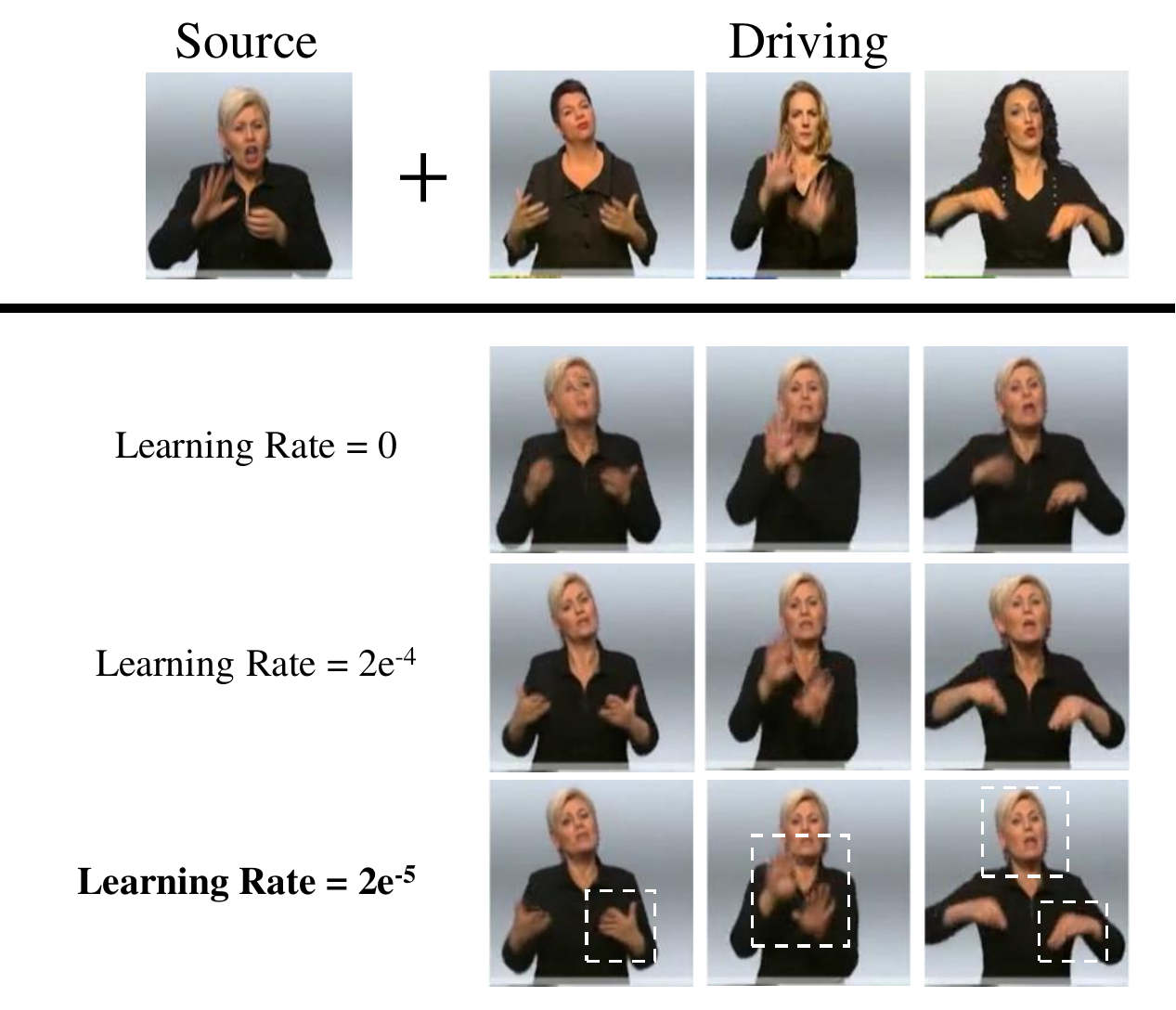}
\end{center}
   \caption{\textbf{Ablation study on different learning rates of the keypoint detector network.} We provide the visual result on the RWTH-PHOENIX-Weather 2014T dataset using different learning rates of the keypoint detector network. The results suggest that fixing the parameters of the keypoint detector network misses finger details and suffers from instability. Meanwhile, a large learning rate could distort the face. Hence, we apply a small learning rate to preserve the finger details while avoiding distortions in the face.}
\label{fig:kplr}
\end{figure}

\textbf{Shall we fine-tune the keypoint detection network?} 
In Section~\ref{KDM}, we mention that the keypoint detector network is jointly trained with the generator network using a small learning rate. To explore whether the fine-tuning procedure of the keypoint detector network is essential, we conduct an ablation experiment on the RWTH-PHOENIX-Weather 2014T dataset. In particular, we set the learning rate of the keypoint detector network to 0, 0.0002, and 0.00002 and test the motion transfer ability of the models. Note that 0 indicates freezing the parameters of the keypoint detector network whereas 0.0002 means setting the learning rate of the keypoint detector network to the same as the learning rate of the other modules. The visual results are shown in Figure~\ref{fig:kplr} and 0.00002 turns out to be the proper learning rate for the keypoint detector network. When we fix the parameters of the keypoint detector network, the transfer results have poor finger details since the output value of the keypoint detector is not precise. The pre-trained keypoint detector model does not leverage the temporal information in video clips, leading to instability when facing blurred frames. In contrast, as shown in Figure~\ref{fig:kplr}, the generator model generates fine-grained finger details when the learning rate of the keypoint detector network is not 0, demonstrating the importance of the fine-tuning process. We can also observe that a large learning rate brings distortions in the face. The reason is that a violent fine-tuning procedure forces the pre-trained model to forget the prior human body structure information, making the generator model capture the wrong identity and background texture information. In other words, the detected keypoints lose their semantics after a drastic fine-tuning process, which could even harm the precision of the keypoint location in some specific body parts such as the face and fingers. Theoretically, if the training process is long enough, the pre-trained keypoint detector model could converge to the same trivial solution as the unsupervised keypoint detectors used in existing methods. Therefore, to keep fine-grained finger details while avoiding losing body structure information, we set the learning rate of the keypoint detector network to 0.00002.

\section{Conclusion} \label{conclusion}
In this paper, we propose a sign language motion transfer framework called Structure-aware Temporal Consistency Network  (STCNet). Different from existing works, STCNet leverages prior human body structure knowledge and the temporal consistency for sign language video generation.
On one hand, we harness a pre-trained keypoint detector network to involve fine-grained human body keypoints in the training process, and fine-tune the keypoint detector network in an end-to-end manner, yielding a more explainable and robust sign language motion.  
On the other hand, we introduce a pair of cycle-consistency losses, \ie short-term cycle loss and long-term cycle loss, to fully exploit temporal information within sign language videos and further improve the temporal continuity of the generated videos. 
Extensive qualitative and quantitative experiments verify that our method could generate competitive videos with accurate motion and high-fidelity video continuity compared with existing works.
We observe that our approach can preserve fine-grained identity attributes of the source image, such as face and hair. 
In the future, we will continue exploring the potential of applying this method to other relevant research fields, such as data augmentation for sign language recognition~\cite{camgoz2018neural, hu2021global, hu2021signbert, tunga2020pose, albanie2020bsl, cui2019deep} and clothing / makeup try-on according to keypoints~\cite{hu2022spg,Han_2018_CVPR,huang2021real}.

\ifCLASSOPTIONcaptionsoff
  \newpage
\fi



\bibliographystyle{IEEEtran}

\end{document}